\documentclass[lettersize,journal]{IEEEtran}
\usepackage{amsmath,amsfonts}
\usepackage{algorithmic}
\usepackage{algorithm}
\usepackage{array}
\usepackage[caption=false,font=normalsize,labelfont=sf,textfont=sf]{subfig}
\usepackage{textcomp}
\usepackage{stfloats}
\usepackage{url}
\usepackage{verbatim}
\usepackage{graphicx}
\usepackage{booktabs}
\usepackage{cite}
\begin{document}
\title{Forecast-aware Gaussian Splatting for Predictive 3D Representation in Language-Guided Pick-and-Place Manipulation}

\author{Kaixin Jia and Jiacheng Xu%
\thanks{Kaixin Jia and Jiacheng Xu are with KTH Royal Institute of Technology, Stockholm, Sweden.}%
\thanks{Emails: Kaixin Jia (kaixinj@kth.se), Jiacheng Xu (jiacxu@kth.se).}%
}



\maketitle
\pagestyle{empty}
\thispagestyle{empty}
\begin{abstract}

We introduce Forecast-aware Gaussian Splatting (Forecast-GS), a predictive 3D representation framework for language-conditioned robotic manipulation. While recent manipulation systems have made progress by grounding language instructions into robot affordances, value maps, or relational keypoint constraints, they usually reason over the current scene and do not explicitly model the task-completed state. This limitation is critical when success depends on satisfying spatial and semantic goals under partial observations, where the robot must evaluate whether a candidate action leads to a feasible task-consistent outcome.

To address this challenge, Forecast-GS augments semantic 3D Gaussian Splatting with task-conditioned final-state prediction. Given multi-view RGB-D observations, our method builds an explicit Gaussian field that encodes scene geometry, appearance, and semantic features distilled from pretrained vision models. A natural language instruction is then decomposed into spatial constraints and grounded within the 3D representation. Forecast-GS generates multiple candidate final states by transforming object-level Gaussian primitives and evaluates these candidates using geometric consistency, collision avoidance, and relational alignment.

We validate Forecast-GS on real-world pick-and-place manipulation tasks, including Cutter-to-Box, Apple-to-Bowl, and Sponge-to-Tray. For each task, we conduct 25 real-world trials under varied initial object configurations using the same robot platform and sensing setup. Forecast-GS with automatic candidate selection achieves success rates of 21/25, 23/25, and 16/25 on the three tasks, respectively, outperforming the ReKep baseline, which achieves 15/25, 19/25, and 10/25. A diagnostic human-assisted setting further improves success rates to 23/25, 24/25, and 19/25, suggesting that candidate generation is effective while automatic ranking remains imperfect. These results suggest that explicitly forecasting task-completed 3D states enables more reliable action evaluation, while the gap between automatic and human-assisted selection indicates that robust final-state ranking remains an important challenge for fully autonomous manipulation. Overall, Forecast-GS provides an interpretable bridge between language understanding, 3D perception, and robotic manipulation planning.
\end{abstract}

\begin{IEEEkeywords}
Robot Manipulation, 3D Scene Representation, Gaussian Splatting, Vision-Language Integration, Predictive Modeling, Task Planning
\end{IEEEkeywords}

\section{Introduction}
\IEEEPARstart{R}{eliable}  robotic manipulation in complex environments requires not only accurate perception of the current scene, but also the ability to anticipate the outcome of actions. For tasks such as object placement, object rearrangement, and region-conditioned pick-and-place manipulation, a robot must evaluate whether a candidate action will produce a feasible final state that satisfies both geometric and semantic constraints. However, most existing approaches plan actions based primarily on current observations or sparse task constraints, without explicitly modeling the post-manipulation scene in 3D. This limitation often leads to unstable or suboptimal behavior in the presence of occlusion, uncertain geometry, or ambiguous language instructions.

Recent advances in vision-language models (VLMs) and large language models (LLMs) have enabled natural language instructions to be translated into robot-executable representations, such as affordances, value maps, and relational constraints~\cite{Ahn2022SayCan,huang2023voxposer,huang2024rekep}. For example, VoxPoser maps language-conditioned objectives into compositional 3D value maps~\cite{huang2023voxposer}, while ReKep formulates manipulation as reasoning over relational keypoint constraints~\cite{huang2024rekep}. Although these approaches improve generalization in language-conditioned manipulation, they typically operate on sparse keypoints or implicit representations, and do not explicitly construct an editable and visualizable 3D representation of the task-completed state.

In parallel, advances in 3D scene representations have significantly improved perception for robotics. Neural Radiance Fields (NeRF) and their language-embedded extensions aggregate geometry and semantics from multi-view observations~\cite{mildenhall2022nerf,kerr2023lerf,shen2023dff,rashid2023lerf_togo,NEURIPS2022_93f25021}. However, these methods often suffer from high optimization cost and limited editability, making them less suitable for interactive manipulation. In contrast, 3D Gaussian Splatting (3DGS) represents scenes using explicit Gaussian primitives, enabling efficient rendering and direct object-level editing~\cite{kerbl3Dgaussians}. Recent works further extend 3DGS with language and self-supervised features, enabling open-vocabulary querying, instance-level segmentation, and robotic grasping~\cite{yu2024languageembeddedgaussiansplatslegs,zhou2024feature3dgssupercharging3d,qiu2024featuresplattinglanguagedrivenphysicsbased,qin2024langsplat3dlanguagegaussian,zheng2024gaussiangrasper}. Despite these advances, existing methods primarily focus on understanding the current scene, and do not explicitly model task-conditioned future states.

In this paper, we propose \textbf{Forecast-aware Gaussian Splatting (Forecast-GS)}, a framework that introduces a \emph{predictive 3D representation} for robotic manipulation. Unlike conventional 3D representations that encode geometry and semantics alone, Forecast-GS augments 3D Gaussian Splatting with task-conditioned final-state prediction, enabling the system to generate, edit, and evaluate multiple candidate manipulation outcomes directly in explicit 3D space. By grounding language-derived spatial constraints in the Gaussian field, our method provides a visualizable and verifiable intermediate representation that bridges perception and planning.

Specifically, we construct a semantically enriched 3D Gaussian field from multi-view RGB-D observations, distilling features from pretrained vision models such as Detic, DINOv2, and CLIP to support open-vocabulary object retrieval and object-level editing~\cite{zhou2022detecting,oquab2023dinov2,radford2021clip}. Given a natural language instruction, spatial constraints are generated and applied within the 3D representation to synthesize candidate final states. These candidates are evaluated using geometric consistency, collision constraints, and relational alignment, providing interpretable goal states for downstream motion planning.

Compared with methods based on 2D features or sparse keypoints, Forecast-GS provides: (1) an explicit and editable 3D task representation; (2) direct prediction and visualization of post-action states; and (3) a unified interface between language understanding, 3D perception, and motion planning. We validate the proposed framework on real-world robotic manipulation tasks and analyze the role of predictive 3D representations in improving task success and execution stability.

The main contributions of this work are summarized as follows:
\begin{itemize}
    \item We introduce a forecast-aware 3D Gaussian representation that explicitly models task-conditioned candidate final states for language-guided pick-and-place manipulation.
    \item We propose a candidate final-state generation and evaluation framework that grounds language-derived spatial constraints in an editable 3D Gaussian field.
    \item We validate the proposed framework on three real-world manipulation tasks and analyze the effects of depth supervision and candidate selection on task execution.
\end{itemize}
\begin{figure}[t]
\centering
\includegraphics[width=\linewidth]{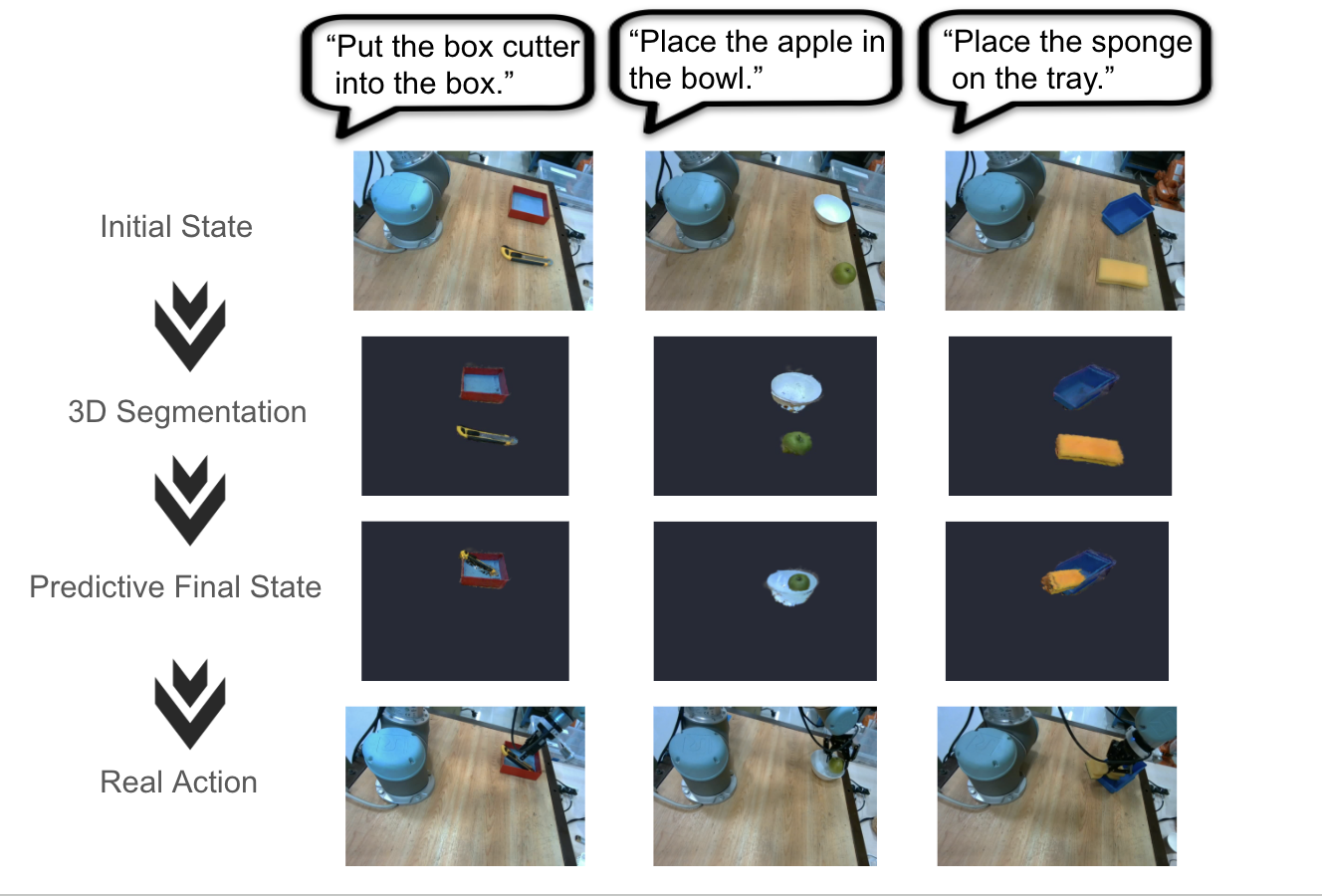}
\caption{
Overview of Forecast-GS. The system integrates language understanding, semantic 3D Gaussian Splatting, and predictive final-state generation. Multiple candidate task outcomes are synthesized and evaluated in 3D before executing the selected manipulation trajectory.
}
\label{fig:overview}
\end{figure}
As illustrated in Fig.~\ref{fig:overview}, our framework explicitly models candidate task outcomes in a unified 3D representation, enabling evaluation prior to execution.

\section{Related Work}

\subsection{Language-Conditioned Robotic Manipulation}

Recent advances in vision-language models (VLMs) and large language models (LLMs) have enabled robots to translate natural language instructions into executable task representations. For example, SayCan grounds language in robot affordances for task decomposition~\cite{Ahn2022SayCan}, VoxPoser constructs compositional 3D value maps for manipulation~\cite{huang2023voxposer}, and ReKep formulates manipulation as spatio-temporal reasoning over relational keypoint constraints~\cite{huang2024rekep}.

While these approaches improve generalization in language-conditioned manipulation, they primarily rely on sparse keypoints, implicit value fields, or constraint optimization in the current state. As a result, they do not explicitly construct or visualize task-completed 3D scenes, making it difficult to evaluate candidate outcomes prior to execution.

Among these methods, ReKep is the closest baseline to our setting because it converts language instructions into relational keypoint constraints and supports real-world manipulation execution. Therefore, we use ReKep as the primary baseline in our experiments. Methods such as SayCan focus more on high-level skill selection, while language-embedded 3D representations mainly address scene understanding rather than task-conditioned final-state prediction.

\subsection{Semantic 3D Scene Representation}

Recent work has explored integrating semantics into 3D scene representations. Neural Radiance Fields (NeRF) and their extensions, such as LERF and DFF, enable joint modeling of geometry and semantics from multi-view observations~\cite{mildenhall2022nerf,kerr2023lerf,shen2023dff,rashid2023lerf_togo,NEURIPS2022_93f25021}. Other approaches incorporate vision-language features for open-vocabulary 3D perception and mapping.

However, these methods are typically based on implicit representations, which are computationally expensive and difficult to edit interactively. Moreover, they mainly focus on understanding the current scene, rather than modeling task-conditioned future states.

\subsection{Gaussian Splatting for Robotics}

3D Gaussian Splatting has recently emerged as an efficient representation for real-time 3D reconstruction and rendering~\cite{kerbl3Dgaussians}. Extensions such as LangSplat, LEGS, and Feature 3DGS incorporate language and self-supervised features into Gaussian fields, enabling open-vocabulary querying and segmentation~\cite{qin2024langsplat3dlanguagegaussian,yu2024languageembeddedgaussiansplatslegs,zhou2024feature3dgssupercharging3d}. GaussianGrasper further applies these representations to robotic grasping~\cite{zheng2024gaussiangrasper}.

Despite these advances, existing works primarily focus on scene understanding and object localization, and do not address task-conditioned prediction of manipulation outcomes.

Our proposed Forecast-GS extends Gaussian representations with predictive modeling, enabling the generation and evaluation of candidate final states.

\section{Method}

\begin{algorithm}[!b]
\caption{Forecast-GS Pipeline}
\label{alg:forecastgs}
\begin{algorithmic}[1]
\REQUIRE Multi-view RGB-D observations $\mathcal{I}$, language instruction $l$
\ENSURE Selected final-state goal $\mathcal{G}^{*}$ for motion planning
\STATE Construct an explicit Gaussian scene representation $\mathcal{S}$ from $\mathcal{I}$
\STATE Distill semantic features from pretrained vision models into Gaussian primitives
\STATE Parse instruction $l$ into a set of spatial constraints $\mathcal{C}$
\STATE Ground object references and spatial relations in the semantic Gaussian field
\STATE Sample candidate object transformations $\{T_k\}_{k=1}^{K}$
\STATE Generate candidate final states $\mathcal{G}_k = T_k(\mathcal{S})$
\STATE Evaluate each candidate using task consistency and physical feasibility scores
\STATE Select $\mathcal{G}^{*} = \arg\min_{\mathcal{G}_k} E(\mathcal{G}_k) + \lambda \Phi(\mathcal{G}_k)$
\STATE Execute a motion plan toward the selected final-state goal
\end{algorithmic}
\end{algorithm}

Given multi-view RGB-D observations $\mathcal{I}$ and a language instruction $l$, Forecast-GS constructs a semantic 3D representation $\mathcal{S}$, generates candidate final states $\{\mathcal{G}_k\}$, and selects a feasible task-consistent goal $\mathcal{G}^{*}$ for motion planning. The framework consists of semantic 3D scene representation, language-conditioned constraint grounding, and predictive final-state generation and evaluation. 

\subsection{Semantic 3D Gaussian Representation}

Given multi-view RGB-D observations $\mathcal{I} = \{I_i, D_i\}_{i=1}^N$, we construct an explicit 3D scene representation using Gaussian Splatting. Each Gaussian primitive is parameterized as:

\begin{equation}
g_j = (\mathbf{x}_j, \Sigma_j, \mathbf{c}_j, \mathbf{f}_j),
\end{equation}

where $\mathbf{x}_j \in \mathbb{R}^3$ denotes the 3D position, $\Sigma_j$ the covariance, $\mathbf{c}_j$ the color, and $\mathbf{f}_j$ a high-dimensional semantic feature.

To incorporate semantic information, we distill features from pretrained vision models (e.g., CLIP, DINOv2) into the Gaussian representation:

\begin{equation}
\mathcal{L}_{feat} = \sum_{j} \| \mathbf{f}_j - \phi(I, \mathbf{x}_j) \|^2,
\end{equation}

where $\phi(\cdot)$ denotes the image feature extractor projected into 3D space.

This results in a unified representation:

\begin{equation}
\mathcal{S} = \{ g_j \}_{j=1}^M,
\end{equation}

which supports both geometric rendering and semantic querying.

\subsection{Language-Conditioned Constraint Grounding}

Given a natural language instruction $l$, we extract a set of spatial constraints:

\begin{equation}
\mathcal{C} = \{ c_k \}_{k=1}^K,
\end{equation}

where each constraint $c_k$ encodes relations such as proximity, containment, or alignment.

These constraints are grounded in the 3D representation by defining a scoring function over object configurations:

\begin{equation}
E(\mathcal{G}) = \sum_{k} w_k \cdot E_k(\mathcal{G}),
\end{equation}

where $\mathcal{G}$ denotes a candidate scene configuration, and $E_k$ measures the violation of constraint $c_k$.

\subsection{Predictive Final-State Generation}

To forecast task outcomes, Forecast-GS generates candidate final states by applying object-level transformations to the semantic Gaussian field. After grounding the manipulated object and target region from the instruction, we extract the corresponding Gaussian primitives using semantic similarity and spatial consistency. In practice, object-level Gaussian subsets are obtained by combining open-vocabulary semantic scores with spatial clustering in the reconstructed workspace.  A candidate transformation $T_k$ is then defined over the selected object-level Gaussian subset, including translation and rotation parameters within the reachable workspace.

For a manipulation task, the transformed scene is represented as:
\begin{equation}
\mathcal{G}_k = T_k(\mathcal{S}),
\end{equation}
where $\mathcal{G}_k$ denotes a possible task-completed scene configuration. Instead of predicting a single action target, Forecast-GS samples translations around the language-grounded target region and rotations around the vertical axis within the reachable workspace. This produces a discrete set of plausible final states, such as alternative placements inside a container or stable poses on a target surface.

The candidate generation process is constrained by the semantic target region and the estimated object geometry. Candidates outside the robot workspace or inconsistent with the target relation are discarded before motion planning. The remaining candidates are visualized in the 3D Gaussian representation and passed to the evaluation module.

\subsection{Candidate Evaluation and Selection}

Each candidate final state is evaluated according to both task consistency and physical feasibility. We define the overall score as:
\begin{equation}
\mathcal{G}^* = \arg\min_{\mathcal{G}_k} \; E(\mathcal{G}_k) + \lambda \Phi(\mathcal{G}_k),
\end{equation}
where $E(\mathcal{G}_k)$ measures violation of language-derived spatial constraints, and $\Phi(\mathcal{G}_k)$ measures physical infeasibility.

The task consistency term is computed as:
\begin{equation}
E(\mathcal{G}_k) = \sum_{m} w_m E_m(\mathcal{G}_k),
\end{equation}
where each $E_m$ corresponds to a grounded spatial relation, such as proximity to a target region, containment inside a container, or alignment with a reference object. The weights $w_m$ control the relative importance of different constraints.

The feasibility term $\Phi(\mathcal{G}_k)$ is computed from three heuristic penalties: collision between the transformed object Gaussians and the remaining scene, violation of the predefined robot workspace bounds, and placement instability estimated from the support relation between the object footprint and the target surface. Candidates with severe workspace or collision violations are discarded before ranking. In practice, these terms provide a ranking over candidate final states before motion planning.

We consider two selection modes in the experiments. In the automatic mode, the final state is selected by the scoring function. In the human-assisted mode, a user selects one final state from the generated candidate set. The human-assisted setting is used only as a diagnostic upper bound for candidate-set quality; it is not part of the autonomous Forecast-GS pipeline. The performance gap between the two modes reflects the remaining challenge of automatic candidate ranking.

\section{Experimental Evaluation and Analysis}

We evaluate the proposed Forecast-GS framework on real-world robotic manipulation tasks to assess its ability to generate and utilize predictive 3D representations for task execution. The experiments are conducted on a UR5 robotic arm equipped with an RGB-D sensor.

\subsection{Experimental Setup}
\begin{figure}[t]
\centering
\includegraphics[width=\linewidth]{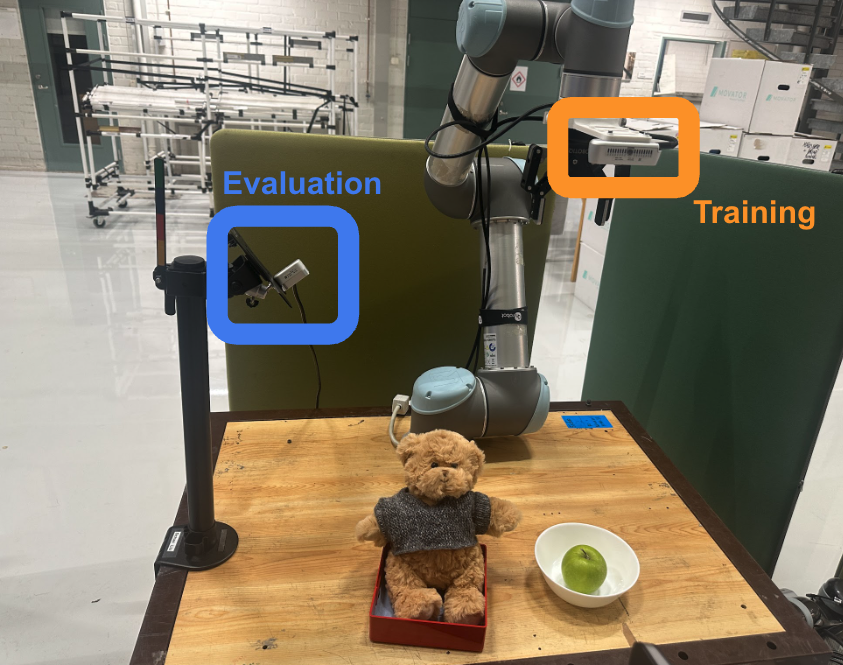}
\caption{
Real-world experimental setup. Multi-view RGB-D cameras are used for 3D reconstruction and forecasting, while a separate observation camera is used for evaluation.
}
\label{fig:fig3}
\end{figure}
We consider three representative pick-and-place tasks: \emph{Cutter-to-Box}, \emph{Apple-to-Bowl}, and \emph{Sponge-to-Tray}. Each task involves spatial reasoning under partial observations and requires the robot to place an object into a target region specified by language instructions. As illustrated in Fig.~\ref{fig:fig3}, the system employs multiple RGB-D cameras for 3D reconstruction and predictive modeling, while a separate observation camera is used for evaluation.

For each task, the system takes multi-view RGB-D observations and a language instruction as input, constructs a semantic 3D Gaussian representation, and generates multiple candidate final states. The selected state is then executed by a motion planner.

\begin{figure*}[!t]
    \centering
    \includegraphics[width=\textwidth]{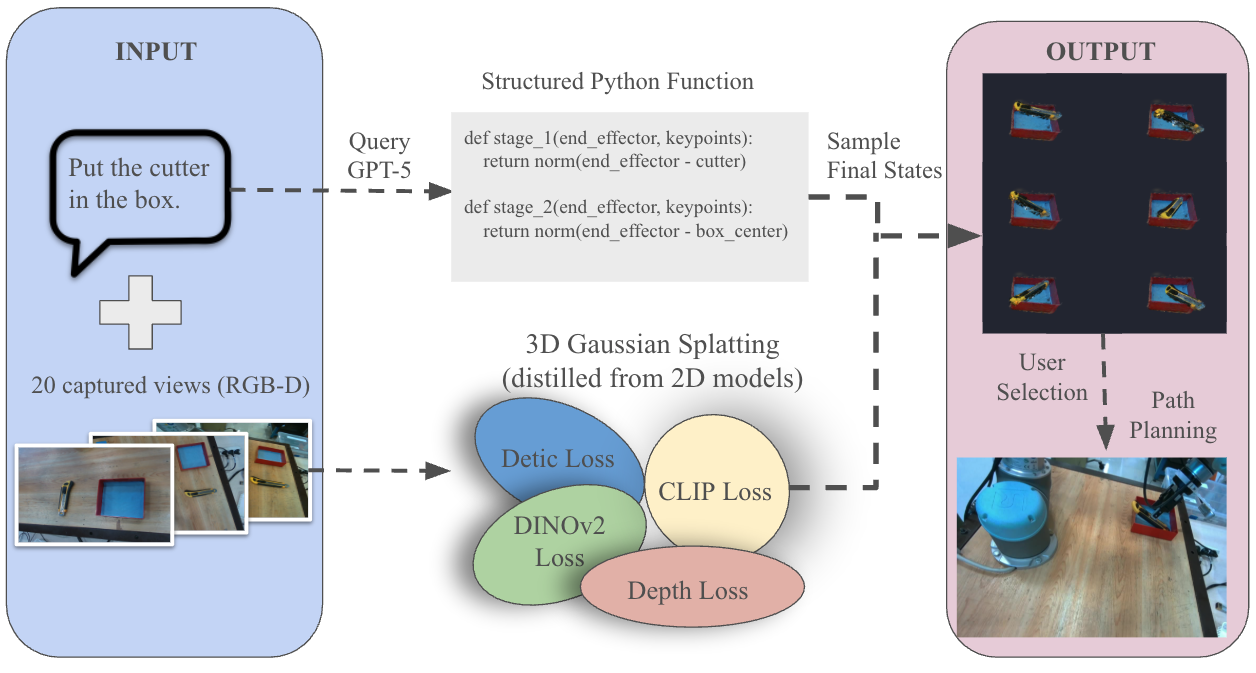}
    \caption{Forecast-GS Pipeline. Given a natural language instruction and multi-view RGB-D observations, the system constructs a semantic 3D Gaussian representation, grounds task constraints in 3D, generates multiple candidate final states, and selects a goal configuration for motion planning. Candidate selection can be performed automatically using the scoring function or manually in the human-assisted setting.}
    \label{fig:pipeline}
\end{figure*}
Fig.~\ref{fig:pipeline} illustrates the end-to-end process of our method. Starting from initial observations, the system reconstructs a semantic 3D scene, predicts multiple candidate final states, and executes the selected action. 

We observe that Forecast-GS is able to generate spatially consistent and semantically meaningful final states, even in the presence of partial observations. The predicted states provide an interpretable intermediate representation, allowing visualization of task outcomes before execution.
\subsection{Evaluation Protocol}

For each task, we conduct 25 real-world trials using the same UR5 robot platform, RGB-D sensing setup, and task instruction format. Each trial starts from a varied initial object configuration within the robot workspace. A trial is considered successful if the manipulated object is placed in the specified target region and remains stable after release. We report successful placement rate, maximum consecutive successes, mean consecutive successes, and final position error when the target position can be measured. Final position error is reported only when the final object position can be reliably measured after execution.

We compare Forecast-GS with ReKep, which is the closest baseline to our setting because it represents language-conditioned manipulation tasks through relational keypoint constraints and supports real-world execution. ReKep is evaluated under the same robot setup, task definitions, language instructions, initial-state distribution, and success criteria. We additionally evaluate ablated variants of Forecast-GS to isolate the effects of depth supervision and candidate final-state selection.

We report both automatic and human-assisted selection results. The automatic setting corresponds to the fully autonomous pipeline, while the human-assisted setting evaluates the quality of the generated candidate set when candidate ranking is not the limiting factor.

\subsection{Quantitative Evaluation}
\begin{table*}[!t]
  \centering
  \footnotesize
  \caption{Real-world manipulation performance across three pick-and-place tasks. Forecast-GS is evaluated without depth supervision, with automatic candidate selection, and with human-assisted candidate selection.}
  \label{tab:perf}
  \setlength{\tabcolsep}{3pt}
  \renewcommand{\arraystretch}{1.1}
  \resizebox{\textwidth}{!}{%
  \begin{tabular}{l ccc c ccc c ccc c}
    \toprule
    & \multicolumn{4}{c||}{\textbf{Cutter to Box}}
    & \multicolumn{4}{c||}{\textbf{Apple to Bowl}}
    & \multicolumn{4}{c}{\textbf{Sponge to Tray}} \\
    \cmidrule(lr){2-5}\cmidrule(lr){6-9}\cmidrule(lr){10-13}
    & \multicolumn{3}{c|}{\textbf{Forecast-GS (ours)}} & \textbf{ReKep}
    & \multicolumn{3}{c|}{\textbf{Forecast-GS (ours)}} & \textbf{ReKep}
    & \multicolumn{3}{c|}{\textbf{Forecast-GS (ours)}} & \textbf{ReKep} \\
    & w/o Depth & Auto Selection & Human-assisted & ReKep
    & w/o Depth & Auto Selection & Human-assisted & ReKep
    & w/o Depth & Auto Selection & Human-assisted & ReKep \\
    \midrule
    Max Consecutive & 0 & 8 & \textbf{11} & 6
                    & 2 & 11 & \textbf{13} & 7
                    & 0 & 6 & \textbf{7}  & 4 \\
    Mean Consecutive & 0 & 6.4 & \textbf{9.6} & 4.2
                     & 0.5 & 9.3 & \textbf{11.8} & 6.5
                     & 0 & 5.3 & \textbf{7.6} & 3.9 \\
    Successful Place Rate & 0/25 & 21/25 & \textbf{23/25} & 15/25
                          & 3/25 & 23/25 & \textbf{24/25} & 19/25
                          & 0/25 & 16/25 & \textbf{19/25} & 10/25 \\
    Mean Position Error (cm) & \textemdash & 1.8 & \textbf{1.5} & \textemdash
                         & 6.1 & 5.6 & \textbf{5.0} & \textemdash
                         & \textemdash & 3.8 & \textbf{3.3} & \textemdash \\

    Std Position Error (cm)  & \textemdash & 0.7 & \textbf{0.6} & \textemdash
                         & 2.5 & 2.1 & \textbf{1.8} & \textemdash
                         & \textemdash & 1.3 & \textbf{1.1} & \textemdash \\
    \bottomrule
  \end{tabular}}
\end{table*}
Table~\ref{tab:perf} reports the real-world manipulation results across the three tasks. Forecast-GS with human-assisted candidate selection achieves the strongest performance, indicating that the generated candidate final states often contain feasible and task-consistent outcomes. The automatic selection variant also achieves higher successful placement rates and longer consecutive success sequences than ReKep across all three tasks, suggesting that explicit final-state prediction provides useful structure for manipulation planning even without manual selection.

The performance gap between automatic and human-assisted selection shows that candidate generation and candidate ranking are two related but distinct challenges. While Forecast-GS can generate high-quality final-state candidates, automatically selecting the optimal candidate remains difficult in ambiguous or cluttered scenes.

We emphasize that the improvement is primarily attributed to the ability to explicitly model and evaluate candidate final states in 3D space, rather than relying solely on intermediate representations defined in the current scene. These results suggest that explicit final-state candidates provide an interpretable intermediate representation for verifying manipulation outcomes before execution.

\subsection{Ablation Study}
To evaluate the contribution of key components in our framework, 
we conduct ablation studies on depth supervision and candidate selection.

Specifically, we consider three configurations:
\begin{enumerate}
    \item w/o Depth, where depth supervision is removed from the Gaussian representation;
    \item Auto Selection, where the final state is selected automatically using the proposed scoring function;
    \item Human-assisted Selection, where a user selects the final state from the generated candidates.
\end{enumerate}

As shown in Table~\ref{tab:perf}, removing depth supervision leads to a significant drop 
in performance (e.g., 0/25 success rate in multiple tasks), indicating that geometric 
consistency is critical for constructing reliable predictive representations.

Comparing Auto Selection with Human-assisted Selection, we observe that automatic candidate ranking remains challenging. Human-assisted selection improves performance because it can choose feasible final states from the generated candidate set, suggesting that the proposed representation often provides valid task outcomes even when the automatic ranking is imperfect.

Overall, these results demonstrate that both accurate 3D geometry and effective 
candidate evaluation are essential for predictive manipulation.

\subsection{Failure Cases}

Despite the overall performance improvements, we observe several failure cases 
that reveal the limitations of the proposed framework.

First, in some scenarios, the system generates candidate states that satisfy 
individual spatial constraints but fail to achieve global task consistency, 
resulting in incorrect object placement.

Second, the evaluation function may fail to fully capture physical feasibility, 
leading to selected states with minor collisions or unstable configurations.

Third, without depth supervision, the system struggles to maintain accurate 
3D geometry, which leads to severe degradation in performance, as shown 
in Table~\ref{tab:perf}.

Finally, automatic candidate selection remains a challenging problem. 
In some cases, the system selects suboptimal states due to imperfect 
constraint weighting or ambiguous language grounding.

These results indicate that improving automatic candidate scoring is a key direction for making Forecast-GS fully autonomous. In the current system, human-assisted selection helps evaluate the quality of the generated final-state candidates, but it does not eliminate the need for stronger automatic ranking in future work.

\section{Limitations}

This work provides an initial real-world validation of forecast-aware 3D representations rather than a comprehensive benchmark across all manipulation categories. The current evaluation is limited to three pick-and-place tasks in relatively structured environments, and extension to long-horizon rearrangement and more complex object interactions remains future work. Candidate generation and evaluation rely on sampling-based transformations and heuristic scoring functions, making automatic final-state selection less reliable than human-assisted selection. Finally, the framework depends on multiple pretrained components and semantic feature distillation, which may introduce sensitivity to perception quality and increase computational cost. Future work will investigate more robust candidate ranking, tighter integration between perception and planning, and more efficient incremental Gaussian updates.

\section{Conclusion}

We presented Forecast-aware Gaussian Splatting (Forecast-GS), a predictive 3D representation framework for language-guided pick-and-place manipulation. By integrating semantic 3D Gaussian representations with language-derived spatial constraints, Forecast-GS generates and evaluates candidate task outcomes directly in 3D before motion planning. Real-world experiments on three pick-and-place tasks show improved performance over a keypoint-based baseline, while ablations demonstrate the importance of depth supervision and candidate selection. These results suggest that explicit final-state prediction in editable 3D representations is a promising direction for more interpretable and reliable robotic manipulation.
\newpage


\bibliographystyle{IEEEtran}
\bibliography{reference}

\end{document}